%% file: main.tex
\pdfoutput=1

\documentclass[11pt]{article}

\usepackage[preprint]{acl}

\usepackage{tabularx}

\usepackage{booktabs}

\usepackage{times}
\usepackage{latexsym}

\usepackage[T1]{fontenc}

\usepackage[utf8]{inputenc}

\usepackage{microtype}

\usepackage{inconsolata}

\usepackage{graphicx}
\usepackage{subcaption}
\usepackage{amsmath}
\usepackage{kotex}
\usepackage{CJKutf8}
\newcommand{\zh}[1]{\begin{CJK*}{UTF8}{gbsn}#1\end{CJK*}}
\usepackage{multirow}
\usepackage{makecell}
\usepackage{xcolor}

\title{AEGIS: Awareness-Enhanced Guidance for Iterative Safeguard}

\author{
 \textbf{Kyungwon Park}\thanks{\scriptsize Dept. of Artificial Intelligence, Yonsei University},
 \textbf{Sangmin Lee}\thanks{\scriptsize Dept. of Electrical \& Electronic Engineering, Yonsei University},
 \textbf{Heejae Chon}\thanks{\scriptsize Dept. of Computer Science \& Engineering, Yonsei University},
 \textbf{Hyungu Kang}\thanks{\scriptsize Dept. of German Language and Literature, Yonsei University}
 \\
 Yonsei University, Seoul, South Korea
 \\
 \small{\texttt{\{cosmicboon, 0914eagle, khgysg012\}@yonsei.ac.kr, sangmin\_lee@dsp.yonsei.ac.kr}}
}
\begin{document}
\maketitle
\thispagestyle{plain}
\pagestyle{plain}
\begin{abstract}
Span-level rationales are often assumed to improve controllability in text detoxification, but it remains unclear when such guidance helps and when it introduces trade-offs. We present \textbf{A}wareness-\textbf{E}nhanced \textbf{G}uidance for \textbf{I}terative \textbf{S}afeguard (\textbf{AEGIS}) as an exploratory framework for studying span-guided multilingual detoxification across English, Mandarin Chinese, and Korean. AEGIS combines span-level detector outputs with frozen generator backbones, allowing harmful spans, intensity labels, and target attributes to be provided as structured guidance during rewriting. Rather than claiming state-of-the-art detoxification performance, we analyze how span guidance affects the balance between toxicity reduction and meaning preservation across generator families, model scales, and languages. Our results suggest that span-guided detoxification is conditionally useful: explicit rationales change the trade-off between toxicity reduction and meaning preservation, but their effects depend strongly on the generator backbone and the linguistic context. These findings highlight both the promise and the limitations of span-level control signals for multilingual detoxification.

\end{abstract}

\section{Introduction}

Recent breakthroughs in large language models (LLMs) have ushered in a new era of AI-powered conversational services, ranging from customer support chatbots and virtual tutors to mental health assistants~\cite{mohapatra2024evaluating, gabriel2024relate}. As these systems increasingly mediate human communication, concerns have grown over their potential misuse for generating or amplifying harmful content. This issue has become even more pressing with the rapid expansion of online platforms, which, while enabling large-scale and instantaneous interactions, have also accelerated the spread of hate speech, discrimination, and profanity. Such expressions can reinforce societal biases and even incite real-world violence.

In response, automated detoxification systems have been proposed to mitigate toxic content. However, existing approaches often overlook linguistic and cultural nuances—particularly in non-English contexts. As \citet{lee2023cultural} highlight, sociocultural context critically shapes how hate is expressed and perceived, underscoring the need for culturally informed and fine-grained mitigation strategies.

\input{figures/main_fig}
To tackle the aforementioned challenges, harmful language detection and mitigation have been explored from multiple directions.
Most detection models~\cite{jahan2023survey} operate at the sentence level, assigning a single coarse label (e.g., \texttt{offensive}, \texttt{hate}, or \texttt{normal}) to an entire comment, even when token-level annotations are available. Such coarse-grained classification often fails to capture fine-grained span-level cues that are crucial for nuanced understanding.
On the other hand, text rewriting and style-transfer approaches~\cite{agarwal2023haterephrase} typically modify the entire sentence, which risks distorting the author’s original tone, intent, or semantics—particularly in context-dependent or subtly toxic expressions.
Recently, \citet{kim2022whyishatespeech} proposed a \emph{masked rationale prediction} task that leverages token-level span annotations to enhance both detection accuracy and interpretability. However, such methods remain restricted to the detection stage, and end-to-end pipelines capable of selectively rewriting only the identified harmful spans are still exceedingly rare.

While span-level rationales are intuitively appealing for detoxification, their practical benefit is not yet fully understood. In particular, explicit rationales may help a generator localize harmful content, but they may also encourage over-editing, deletion, or meaning drift when the generator interprets the guidance too aggressively. This paper therefore does not aim to establish a new state-of-the-art detoxification system. Instead, we ask a more diagnostic question: \emph{when does span-level guidance help multilingual detoxification, and when does it hurt?}

We study this question using \textbf{A}wareness-\textbf{E}nhanced \textbf{G}uidance for \textbf{I}terative \textbf{S}afeguard (\textbf{AEGIS}), a modular framework that separates span detection from generation. The detector identifies harmful spans and their intensity, and a separate target classifier predicts protected-group attributes. These signals are then provided to a frozen generator through span- and attribute-conditioned prompts. This design lets us treat the detector as a fixed instrument and examine how different generator backbones respond to the same span-level guidance.

Our empirical focus is exploratory rather than benchmark-driven. We compare generator backbones and model sizes across English, Mandarin Chinese, and Korean, and analyze the resulting trade-off between toxicity reduction and semantic preservation. We also compare guided generation with unguided and reflection-based variants to understand whether explicit span rationales consistently improve the detoxification process.

We summarize our contributions as follows:

\begin{itemize}
  \item We present an exploratory analysis of span-guided multilingual detoxification, focusing on the conditions under which explicit span rationales help or hurt rewriting quality.

  \item We use AEGIS as a modular analytical framework that keeps the span detector fixed while varying frozen generator backbones, enabling controlled comparison of toxicity--meaning trade-offs.

  \item We clarify the limitations of span-guided detoxification, showing that explicit guidance should be viewed as a controllability mechanism rather than a guarantee of superior detoxification performance.
\end{itemize}

\section{Related Work}
\subsection{Hate Speech Detection}
While fine-tuned Transformer-based models have shown strong performance on hate speech detection benchmarks~\cite{jahan2023survey}, recent research has increasingly emphasized interpretability, robustness, and cross-lingual generalization.
Frameworks such as NACL~\cite{masud2022nacl} jointly model hate spans and toxicity intensity,
while recent span-based work on implicit harmful content~\cite{jafari2024targetspan} highlights the role of target spans in recognizing subtler harmful expressions,
enhancing both explainability and boundary consistency.

However, most of these studies remain limited to English, and their effectiveness often degrades when applied to typologically distant or culturally specific languages.
Although several non-English corpora have been introduced—such as KOLD~\cite{jeong2022kold},
K-HATERS~\cite{park2023khaters}, BEEP!~\cite{moon2020beep}, and K-MHaS~\cite{lee2022kmhas} for Korean—
comparatively little work has examined span-level and culturally grounded signals as control inputs for multilingual detoxification.
This gap motivates our multilingual detector, which incorporates language-specific intensity schemes and target prediction into multilingual encoder backbones,
while maintaining sensitivity to local linguistic and cultural nuances.

\input{figures/submodules}
\subsection{Hate Speech Mitigation}
Hate speech mitigation aims to neutralize or rewrite offensive content while preserving the original meaning and intent,
and is commonly framed as a text style transfer problem where toxicity defines the stylistic dimension~\cite{toshevska2021review, jin2022deep}.
Early approaches focused primarily on English datasets, emphasizing surface-level rewriting or lexicon-based substitution.
Recent large language model (LLM)–based methods~\cite{agarwal2023haterephrase} suggest that prompt-driven rephrasings can reduce hate intensity while preserving meaning, yet they may still struggle with subtle, context-dependent, or culturally specific expressions.

Building on these advances, frameworks such as NACL~\cite{masud2022nacl} introduced span-conditioned rewriting with adversarial feedback,
showing that incorporating structured signals—such as hate spans and severity levels—improves mitigation controllability.
However, most prior systems remain monolingual, often limited to English, and lack cross-lingual adaptability.
This gap is increasingly critical as hate speech manifests differently across languages and sociocultural contexts~\cite{lee2023cultural, jahan2023survey}.
Recent multilingual efforts, including cross-lingual transfer and toxicity translation benchmarks,
highlight the need for frameworks that can reason over culturally grounded hate expressions rather than relying solely on lexical overlap.

To address these challenges, our work explores a modular, multilingual approach that integrates fine-grained span detection and rationale-guided generation,
enabling controlled and culturally aware detoxification across English, Mandarin Chinese, and Korean.

\section{AEGIS as an Exploratory Framework}
AEGIS comprises two core modules: a span-level detector and a rationale-guided generator. In this preprint, we use AEGIS primarily as an analytical framework rather than as a claim of state-of-the-art detoxification performance. The detector provides structured span guidance, and the generator uses that guidance to rewrite harmful content. In the generation experiments, we keep the detector component fixed when comparing frozen generator backbones, so that the reported differences can be interpreted as differences in how generators respond to the same type of span-level guidance.

\subsection{Detector}

\label{sec:detector}
The detector module operates in three sequential stages. We use the same detector architecture across languages but train a separate detector for English, Mandarin Chinese, and Korean because the datasets use different sentence-level label spaces and exhibit distinct linguistic structures. First, in the intensity-aware BIO tagging stage, the input text is tokenized and character-level harmful spans are converted into token-level BIO tags, while target attributes are modeled through a separate classification head. Second, during the multitask fine-tuning stage, a multilingual encoder is optimized through three parallel tasks to perform the desired detection objectives. Finally, in the output span integration stage, the raw BIO tag sequence is transformed into coherent spans, and unreliable detections are filtered out to ensure the quality of the extracted rationales.

\input{tables/tagging}
\vspace{2pt}
\noindent\textbf{Intensity-Aware BIO Tagging.}
Unlike prior approaches that use binary span labeling, our token-level tagging scheme distinguishes between mild and strong offensive spans. We use a 5-way BIO label set: \texttt{O}, \texttt{B-SOFT}, \texttt{I-SOFT}, \texttt{B-HARD}, and \texttt{I-HARD}. Here, soft and hard labels correspond to lower- and higher-intensity harmful spans, respectively. BIO labels are assigned to token/subword units in English and character-level units in Korean and Chinese, but adjacent units are merged into continuous semantic spans; therefore, multi-token or multi-character expressions are not treated as independent keyword matches. Target attributes are handled by a separate multi-label target classification head rather than by additional BIO target tags. This separation avoids conflating the span boundary task with target-group identification while still allowing target information to be supplied to the generator. The label criteria are summarized in Table~\ref{tab:labels}.

\vspace{2pt}
\noindent\textbf{Multitask Fine-Tuning/Inference.}
We employ multilingual encoder backbones, XLM-R and InfoXLM, with three task-specific heads---sentence classification, BIO token labeling, and target attribute classification---to produce per-token and per-sentence predictions. The sentence classification head predicts the dataset-specific toxicity class (3-way for English and Chinese; 4-way for Korean). The BIO token labeling head predicts one of five token-level BIO labels for offensive-span extraction; to enforce valid tag transitions (e.g., preventing an \texttt{I}-tag from starting a span) and reduce boundary errors, we apply a Conditional Random Field (CRF) decoding layer immediately after the softmax. Finally, the target attribute head performs multi-label classification to identify which protected-group targets, if any, are mentioned in the text.

\vspace{2pt}
\noindent\textbf{Output Span Integration.}
Then we convert the raw BIO tag sequence into coherent spans and filter out unreliable detections. First, we merge adjacent units that form a valid BIO sequence with the same intensity into a single span. Next, we discard any span shorter than two tokens—since one‐token spans tend to be noisy—and any span whose maximum softmax confidence (over its constituent tokens) falls below a threshold \(\theta_{\mathrm{span}}=0.6\), as determined on the validation set. The remaining spans constitute our final set of token‐level rationales. We then output this span list along with the overall sentence classification label and the predicted target attributes.

\subsection{Generator}
\label{sec:generator}
The generator module receives the original sentence along with the harmful-span rationales extracted by the detector and iteratively produces a refined, non-toxic rewrite. The workflow proceeds as follows. First, rationale-guided prompting constructs an input prompt for the generator based on the detected spans. Next, an initial mitigation step reduces the toxicity of the input text. Finally, a self-reflection and iterative refinement process is applied to further detoxify the text while preserving its original semantics.

\vspace{2pt}
\noindent\textbf{Generator Models.}
We adopt two open-source LLM families of varying model sizes, LLaMA 3.2 and Qwen 3, as generators without additional fine-tuning; the parameter counts are reported in Table~\ref{tab:main}. We intentionally focus on relatively small generators to better isolate the effect of explicit span guidance, since stronger general-purpose LLMs may partially mask this effect through their built-in detoxification capabilities. We use frozen generators because our goal is to analyze the effect of explicit span guidance rather than to introduce a newly trained detoxification model. This setup also reflects practical scenarios where a fixed multilingual LLM is prompted with structured control signals instead of being fine-tuned for every language or domain.

\input{tables/rationale_prompt}
\vspace{2pt}
\noindent\textbf{Rationale-Guided Prompting.}
After selecting the generator models, we constructed a prompt template to encode each detected span—including its offset, category, target, and intensity—into a natural language instruction. This prompt guides the generator on how to handle lower-intensity versus higher-intensity harmful spans. As illustrated in Table~\ref{tab:rationale}, the template provides fine-grained guidance to the generator module, helping to ensure that the original meaning of the text is preserved during detoxification.

\vspace{2pt}
\noindent\textbf{Initial Mitigation Generation.}
Subsequently, we feed the generator model with the rationale-guided prompt template described above to produce the first detoxified output, referred to as the initial mitigation. Specifically, we employ nucleus sampling with
top\_p value of 0.9 and a temperature of 0.7, generating a maximum of 150 new tokens to obtain the first output $\hat y_1$.

\input{tables/iterative_prompt}
\vspace{2pt}
\noindent\textbf{Self-Reflection and Iterative Refinement.}
After obtaining the first detoxified output $\hat y_1$, it is evaluated by the detector module (we call the critic model), which compares the generated text against the original sentence using automatic metrics such as BERTScore and toxicity. The model then determines whether it satisfies predefined thresholds $(\theta_\text{toxicity}, \sigma_\text{fidelity})$. If any metric falls below its corresponding threshold, the generator is prompted to critique and revise its output using a structured self-reflection prompt, as illustrated in Table~\ref{tab:iterative}. This process is repeated for up to three iterations or until all metrics are satisfied. Once the output $\hat y_t$ meets all criteria, it is emitted as the mitigated sentence $\hat y_{out}$. Optionally, simple post-processing heuristics, such as residual slur removal, are applied to finalize the output.

\section{Experimental Setup}
\label{sec:setup}
\subsection{Datasets}
\noindent\textbf{English.}
HateXplain~\cite{mathew2021hatexplain} is a large-scale English hate speech corpus
containing approximately 19{,}916 posts collected from \textit{Twitter} and \textit{Gab}.
Each post is annotated by three crowdworkers from Amazon Mechanical Turk (AMT)
with three complementary components: (1) a class label
(\textit{Hate Speech}, \textit{Offensive}, or \textit{Normal}),
(2) a target group category (\textit{Religion, Gender, Race, Political, Other}),
and (3) a rationale span that highlights the specific tokens considered hateful.
The dataset achieves an inter-annotator agreement (IAA) of 0.61,
indicating moderate consensus.

\vspace{2pt}
\noindent\textbf{Mandarin Chinese.}
STATE ToxiCN~\cite{bai2025statetoxicn} is a Mandarin Chinese dataset
containing 8{,}029 social media posts, of which 4{,}942
(61.55\%) include hate-related expressions.
It further defines 9{,}533 quadruples
(Target–Argument–Hateful–Group), including 6{,}034
hate-related tuples (63.60\%).
The dataset covers major target types such as \textit{Gender, Region, Race},
and includes 854 ``multi-group'' cases (8.96\%).
A lexicon of 830 hate slang entries—collected and manually interpreted
from Chinese online forums—provides additional contextual cues.

\vspace{2pt}
\noindent\textbf{Korean.}
K-HATERS~\cite{park2023khaters} is a span-level Korean hate speech corpus
comprising 17{,}601 online comments annotated with both
offensiveness and target attributes.
Each comment includes intensity labels
(\textit{Normal, Offensive, L1\_Hate, L2\_Hate})
and rationale spans marking explicit and implicit hate expressions.

\subsection{Experiment Configurations.}
\noindent\textbf{Environments.}
All training of AEGIS was conducted on a single NVIDIA RTX 3090 GPU with 24GB VRAM using mixed-precision (FP16) optimization.
Inference and generation were performed on the two RTX A5000 GPUs with 24GB VRAM across all three languages (English, Korean, and Mandarin Chinese).

\vspace{3pt}
\noindent\textbf{Hyperparameters.}
We used the AdamW optimizer with a learning rate of $2\times10^{-5}$ and a weight decay of 0.01.
The detector module was fine-tuned for 3 or 4 epochs, depending on language (3 epochs for English and Korean, 4 for Mandarin Chinese), with a batch size of 8, and early stopping was applied based on the validation span-F1 score with a patience value of 3.

\vspace{2pt}
\noindent\textbf{Loss Functions.}
The detector of AEGIS employs three prediction heads:
(1) a sentence-level classification head (3-way for English and Chinese, 4-way for Korean),
(2) a 5-way token-level BIO labeling head, and
(3) a multi-label target classification head.
The module is optimized using a combined loss that integrates the objectives of all three tasks, formulated as:
\[
  \mathcal{L} =
  \mathcal{L}_{\mathrm{cls}}
  + \beta \mathcal{L}_{\mathrm{bio}}
  + \mathcal{L}_{\mathrm{target}},
\]
where $\mathcal{L}_{\mathrm{cls}}$ denotes the cross-entropy loss for sentence classification,
$\mathcal{L}_{\mathrm{bio}}$ is the weighted cross-entropy loss for token tagging,
and $\mathcal{L}_{\mathrm{target}}$ represents the binary cross-entropy with logits for target detection. Padding positions are ignored during loss computation; the \texttt{O} class is retained with weight 1.0 as shown in Table~\ref{tab:bio_weights}.

\input{tables/bio_weights}
Specifically, the BIO loss is scaled by a factor of $\beta = 2.0$ to mitigate the severe class imbalance in sequence labeling,
where non-\texttt{O} tokens account for less than 1\% of all tokens.
We then apply language-specific class weights for BIO tagging as specified in Table~\ref{tab:bio_weights}.
These class weights emphasize rare but critical non-\texttt{O} tokens while maintaining training stability.
This approach effectively handles the extreme imbalance between \texttt{O} tokens (over 99\%) and non-\texttt{O} tokens (below 1\%)
without requiring any additional architectural components.

\vspace{-10pt}
\subsection{Evaluation Metrics}
\noindent\textbf{BERTScore.}
BERTScore leverages embeddings from a pretrained BERT model to measure semantic similarity between generated and reference texts. Cosine similarities between embeddings are used to compute precision, recall, and F1 scores, capturing contextual and semantic information beyond exact word matches. Higher scores indicate stronger semantic alignment.

\vspace{2pt}
\noindent\textbf{Toxicity.}
Toxicity measures the extent to which a model output contains inappropriate, offensive, or biased content. Scores are typically computed using specialized detection models and are employed to assess the effectiveness of generation and mitigation strategies. In our experiments, we use both the Perspective API~\cite{Goel2022NewGeneration} and our detector’s span-level scores to calculate toxicity.

\vspace{2pt}
\noindent\textbf{LLM Evaluation.}
We additionally conducted an LLM-based evaluation on two criteria: semantic similarity and toxicity reduction.
Both were rated on a 1–5 scale using Gemini 2.5 Pro~\cite{comanici2025gemini}, where 1 indicates severe semantic distortion or no toxicity reduction, and 5 indicates strong semantic preservation or substantial toxicity reduction.
These LLM judgments complement automatic metrics by providing an additional multilingual assessment signal.
Details of the evaluation prompts and criteria provided to the LLM judge are presented in Table~\ref{tab:evaluation_guidelines}.

\input{tables/guidelines}
\section{Experiments}
In this section, we use AEGIS to examine span-guided detoxification as an exploratory setting. We first present the main detoxification results and interpret them as toxicity--meaning trade-offs rather than as a state-of-the-art benchmark. We then analyze guided generation and iterative refinement, report detector validation as a span-extraction instrumentation check, and discuss qualitative examples.

\subsection{Main Detoxification Results}
\input{tables/aegis_main}

\noindent\textbf{Trade-off with English baselines.}
Table~\ref{tab:main} compares AEGIS variants with ParaDetox~\cite{logacheva2022paradetox} and DetoxLLM~\cite{khondaker2024detoxllm}. Because these baselines are used here only for English detoxification, we report them only for English and leave the Chinese and Korean entries blank. The results should not be interpreted as showing that AEGIS uniformly outperforms prior detoxification systems. Instead, they show that span-guided generation produces a different trade-off profile: some AEGIS variants obtain competitive BERTScore or detector-based toxicity values, while the English baselines remain stronger on other LLM-based scores.

\vspace{2pt}
\noindent\textbf{Generator backbone effects.}
Within the AEGIS pipeline, LLaMA 3.2 and Qwen 3 denote frozen generator backbones rather than separate detector models or separate detector variants. Across languages, we observe that the preferred backbone depends on the evaluation dimension. LLaMA variants tend to preserve semantic similarity more strongly in English, whereas Qwen variants obtain stronger LLM-based toxicity-reduction scores in some Chinese and Korean settings. These patterns support our main interpretation that span guidance has conditional effects, but its effect depends on the generator and language rather than being uniformly beneficial.

\subsection{Guided Generation and Iterative Refinement}
\input{tables/guide_and_reflect}
\input{figures/abl_iteration}

\noindent\textbf{Impact of Guidance Strategy.}
Table~\ref{tab:methods} reports an ablation comparing unguided generation, span-guided generation, and guided generation with reflection. In this experiment, adding span guidance improves BERTScore and LLM-based semantic similarity relative to the unguided setting, while reflection further reduces toxicity but slightly lowers BERTScore. This suggests that guidance and reflection should be treated as controllability mechanisms whose utility depends on the desired balance between toxicity reduction and meaning preservation.

\vspace{2pt}
\noindent\textbf{Impact of Iterative Guidance.}
Figure~\ref{fig:detox_comparison} illustrates how detoxification performance varies with the number of self-reflection iterations. We observe that additional iterations can improve toxicity-related scores, especially in non-English settings, but this improvement may come with a decline in semantic similarity. Rather than indicating that more iterations are always better, the figure illustrates the need to tune the strength of refinement according to task-specific requirements.

\subsection{Detector Validation as Fixed Instrument}
\input{tables/detector}

Table~\ref{tab:detector_performance} presents the performance of the detector used to produce span guidance across English, Korean, and Mandarin Chinese datasets. In this paper, the detector is not the primary object of comparison; it serves as a fixed instrument for generating structured span rationales. Across all three languages, both \textit{XLM-R} and \textit{InfoXLM} achieve high BIO F1 scores in our experiments, suggesting that the span-extraction component provides a usable basis for the exploratory generation analysis.

\vspace{2pt}
\noindent\textbf{Cross-Model Comparison.}
Between the two backbones, \textit{XLM-R} slightly outperforms \textit{InfoXLM} on most metrics, particularly in Non-O F1 and Intensity F1, while \textit{InfoXLM} shows marginally higher Target F1 in English and Korean. These differences indicate that the choice of detector backbone may affect the quality of span guidance; in this preprint, we report these results as detector checks rather than as the main contribution.

\vspace{2pt}
\noindent\textbf{Cross-Lingual Trends.}
English detectors maintain stable Target F1 around 60\%, Korean detectors achieve the highest BIO F1, and Mandarin Chinese models exhibit the strongest Intensity F1. These differences likely reflect annotation schemes and language-specific properties of the datasets. We therefore treat the detector results as instrumentation checks rather than as evidence of broad cross-lingual superiority.

\subsection{Qualitative Analysis}
\input{tables/qualitative_results}

Table~\ref{tab:qualitative} presents qualitative examples of AEGIS outputs in English, Chinese, and Korean. In English, the profanity ``f**king idiot'' is softened to ``really thoughtless.'' In Chinese, a discriminatory phrase describing disabled people as ``a burden to society'' is rewritten as ``a part of society.'' In Korean, the high-intensity phrase ``좌빨 놈들'' is removed while the surrounding proposition is retained. These examples illustrate the intended behavior of span-guided rewriting, but they should be interpreted as illustrative cases rather than evidence that AEGIS consistently preserves all pragmatic or ideological content.

\section{Conclusion}
\label{sec:conclusion}
We presented AEGIS as an exploratory framework for studying span-guided multilingual detoxification across English, Mandarin Chinese, and Korean. Rather than claiming state-of-the-art detoxification performance, our goal was to examine how explicit span-level rationales influence the trade-off between toxicity reduction and meaning preservation. The results suggest that span guidance changes the toxicity--meaning trade-off, and that its effects are conditional on the generator backbone, model scale, language, and evaluation criterion. These findings argue for a cautious view of span-level control: rationales can guide frozen generators, but they do not by themselves guarantee better detoxification. Future work should evaluate stronger and more recent baselines, larger model families, and human judgments to better characterize when span-guided detoxification is preferable to unguided or proprietary LLM-based rewriting.

\clearpage

\appendix
\section{Limitations}
This preprint should be interpreted as exploratory rather than conclusive. First, our baseline coverage is limited. A comprehensive detoxification benchmark would require substantially more recent baselines, more datasets, and stronger proprietary or open-weight LLMs. Second, the generator models used here are relatively small compared to stronger general-purpose LLMs; therefore, the results do not establish that span guidance is necessary or superior when stronger general-purpose models are available. Third, the evaluation relies primarily on automatic metrics and LLM-based judgments. Larger-scale human evaluation is needed to directly assess meaning preservation, over-sanitization, cultural appropriateness, and annotator disagreement. Fourth, the detector depends on supervised span annotations, limiting scalability to low-resource languages without labeled data. Finally, the modular pipeline introduces additional inference cost and may be less suitable for high-throughput moderation systems than simpler end-to-end prompting approaches.

\section{Ethical Statements}
This work uses publicly available datasets that are reported by their providers as anonymized and containing no personally identifying information. We acknowledge that annotator bias may affect the labeling of “intensity” levels; to mitigate class imbalance in detector training, we apply class weighting and validation-based model selection.

Our proposed pipeline—especially the generation module—poses a risk of misuse: a bad actor could invert its behavior to amplify or hallucinate harmful content. We therefore encourage research use with appropriate safeguards and deployment only behind human-in-the-loop moderation interfaces.

We further note that our intensity-aware tagging and guided generation strategies may perform differently across social groups or dialects, potentially leading to over- or under-correction for certain communities. Prior to multilingual or cross-domain adaptation, additional fairness audits and user studies should be conducted.

Finally, before any real-world deployment, we advocate continuous monitoring of false positive/negative cases, routine recalibration of toxicity thresholds, and adherence to responsible AI principles. Our goal is to enable safer online communication while minimizing unintended harms.

\clearpage
\bibliography{custom}

\end{document}

%% file: figures/main_fig.tex
\begin{figure*}[!t]
  \centering
  \includegraphics[width=1.0\linewidth]{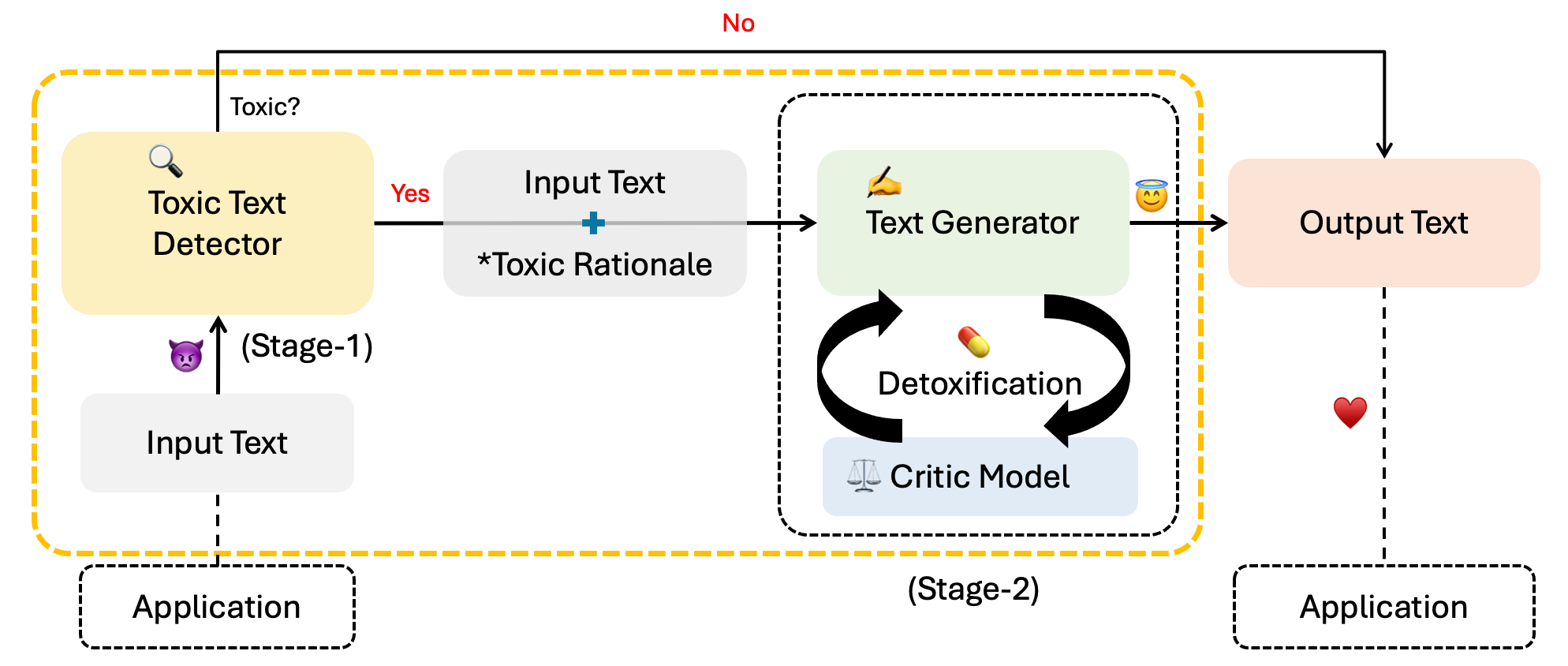}
  \caption{Overall framework of AEGIS.}
  \label{fig:overview}
\end{figure*}

%% file: figures/submodules.tex
\begin{figure*}[!t]
  \centering
  \begin{subfigure}[b]{0.48\linewidth}
    \centering
    \includegraphics[width=\linewidth]{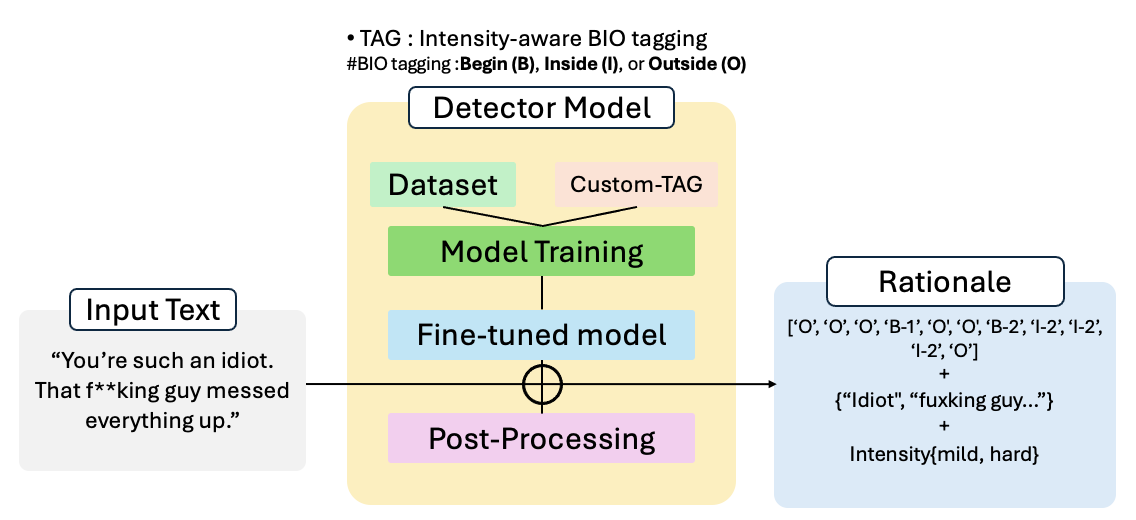}
    \caption{Detector structure}
    \label{fig:detector}
  \end{subfigure}
  \hfill
  \begin{subfigure}[b]{0.48\linewidth}
    \centering
    \includegraphics[width=\linewidth]{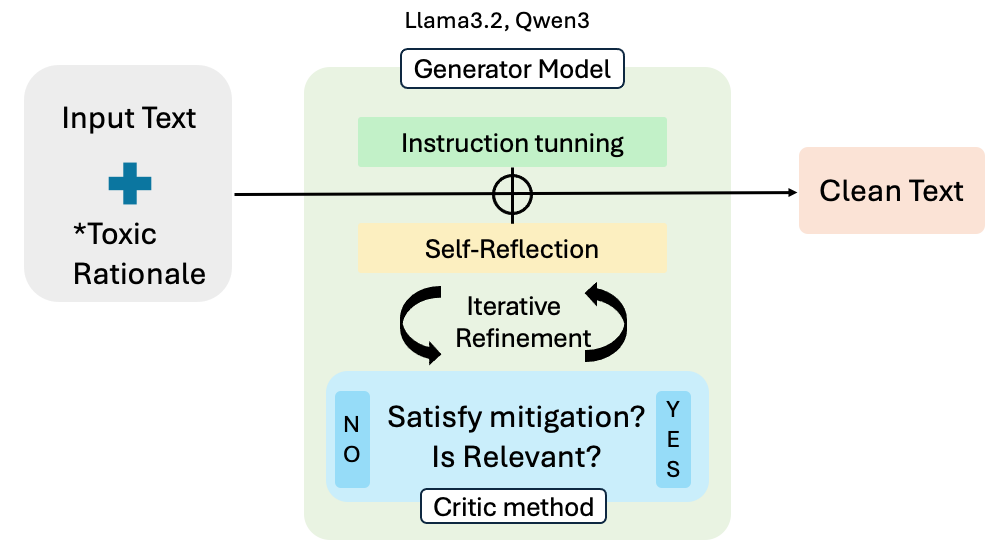}
    \caption{Generator structure}
    \label{fig:generator}
  \end{subfigure}
  \caption{Specific architecture and flow diagram of the detector and generator.}
  \label{fig:detector_generator}
\end{figure*}

%% file: tables/tagging.tex
\begin{table}[!t]
\centering
\resizebox{\columnwidth}{!}{
\begin{tabular}{l|l} \toprule
\textbf{Label} & \textbf{Description} \\ \midrule\midrule
O & Outside of the offensive span (non-offensive) \\
B-SOFT & Beginning of a lower-intensity harmful span \\
I-SOFT & Inside of a lower-intensity harmful span \\
B-HARD & Beginning of a higher-intensity harmful span \\
I-HARD & Inside of a higher-intensity harmful span \\ \bottomrule
\end{tabular}
}
\caption{Five-way token-level BIO labels used for offensive-span extraction. Target attributes are predicted by a separate multi-label head.}
\label{tab:labels}
\vspace{-7pt}
\end{table}

%% file: tables/rationale_prompt.tex
\begin{table}[!t]
\centering
\resizebox{\columnwidth}{!}{
\begin{tabular}{||l||}\toprule
\textbf{Prompt Template} \\ \midrule
Original: \{original\_sentence\}

The following spans are harmful (span, category, target, intensity): \\
\{
  span\_1: “...”, category: offensive, target: none, intensity: SOFT; \\
  span\_2: “...”, category: hate, target: immigrants, intensity: HARD
\} \\
For lower-intensity spans (SOFT), paraphrase gently; \\
For higher-intensity spans (HARD), delete or replace more strongly. \\
Rewrite the entire sentence to be clear and non-toxic: \\
“\{original\_sentence\}” \\ \bottomrule
\end{tabular}
}
\caption{Prompt for initial mitigation.}
\label{tab:rationale}
\end{table}

%% file: tables/iterative_prompt.tex
\begin{table}[!t]
\centering
\small
\resizebox{\columnwidth}{!}{
\begin{tabular}{||l||}\toprule
\textbf{Prompt} \\ \midrule
Generated: \{$\hat y_t$\}\\ \\
Considering the following criteria:\\
1. Has all toxicity been removed?\\
2. Is the original meaning preserved?\\ \\
Provide feedback and rewrite accordingly to produce a new output \{$\hat y_{t+1}$\}. \\ \bottomrule
\end{tabular}
}
\caption{Prompt for iterative refinement.}
\label{tab:iterative}
\vspace{-7pt}
\end{table}

%% file: tables/bio_weights.tex
\begin{table}[!t]
\centering
\resizebox{\columnwidth}{!}{
\begin{tabular}{l|ccccc} \toprule
\textbf{Language} & \textbf{O} & \textbf{B-SOFT} & \textbf{I-SOFT} & \textbf{B-HARD} & \textbf{I-HARD} \\
\midrule\midrule
English & 1.0 & 10.0 & 10.0 & 20.0 & 20.0 \\
Korean  & 1.0 & 8.0  & 4.0  & 15.0 & 8.0  \\
Chinese & 1.0 & 8.0  & 6.0  & 15.0 & 12.0 \\
\bottomrule
\end{tabular}
}
\caption{Token-level class weights for BIO tagging across languages.}
\label{tab:bio_weights}
\vspace{-7pt}
\end{table}

%% file: tables/guidelines.tex
\begin{table}[!t]
\centering
\resizebox{\columnwidth}{!}{
\begin{tabular}{||l||} \toprule
\textbf{Evaluation Guidelines} \\ \midrule
Compare the two sentences below and assign a score. \\[4pt]
\textbf{Original Text (Input Text):} \{input\_text\} \\
\textbf{Model Output (Text):} \{text\} \\[6pt]
\textbf{1. Semantic Similarity (1–5)} \\[2pt]
5: Meaning is almost fully preserved. \\
4: Most of the meaning is preserved. \\
3: Some of the meaning is preserved. \\
2: Meaning is largely changed. \\
1: Meaning is almost completely changed. \\[6pt]
\textbf{2. Toxicity Reduction (1–5)} \\[2pt]
5: Offensive expressions are almost fully removed. \\
4: Most offensive expressions are removed. \\
3: Some offensive expressions are removed. \\
2: Hardly any offensive expressions are removed. \\
1: Offensive expressions mostly remain. \\[4pt]
Assign a score from 1 to 5 for each category. \\
\bottomrule
\end{tabular}
}
\caption{Evaluation prompt provided to the LLM judge for semantic similarity and toxicity reduction scoring.}
\label{tab:evaluation_guidelines}
\end{table}

%% file: tables/aegis_main.tex
\begin{table*}[!t]
\centering
\resizebox{\textwidth}{!}{
\begin{tabular}{l|c|cccc|cccc|cccc} \toprule
\multirow{2}{*}{\textbf{Model}} & \multirow{2}{*}{\textbf{Size}}
& \multicolumn{4}{c}{\textbf{English}}
& \multicolumn{4}{c}{\textbf{Chinese}}
& \multicolumn{4}{c}{\textbf{Korean}} \\
 & & \textbf{BERT} $\uparrow$ & \textbf{Tox.} $\downarrow$ & \textbf{LLM-S} $\uparrow$ & \textbf{LLM-T} $\uparrow$
 & \textbf{BERT} $\uparrow$ & \textbf{Tox.} $\downarrow$ & \textbf{LLM-S} $\uparrow$ & \textbf{LLM-T} $\uparrow$
 & \textbf{BERT} $\uparrow$ & \textbf{Tox.} $\downarrow$ & \textbf{LLM-S} $\uparrow$ & \textbf{LLM-T} $\uparrow$\\
\midrule\midrule
ParaDetox & 0.14B & \textbf{0.91} & \underline{0.34} & \textbf{5.00} & \underline{1.53} & - & - & - & - & - & - & - & - \\
DetoxLLM & 7B & 0.83 & 0.40 & 3.60 & \textbf{4.90} & - & - & - & - & - & - & - & - \\ \midrule
\multirow{2}{*}{\makecell[l]{AEGIS \\ (LLaMA 3.2 generator)}}
 & 1.3B & \underline{0.89} & \underline{0.34} & 3.00 & 4.00 & \underline{0.81} & \textbf{0.34} & 3.25 & 3.06 & \textbf{0.89} & \textbf{0.34} & 2.45 & 2.09 \\
 & 3.3B & \textbf{0.91} & \textbf{0.32} & 3.03 & 4.43 & \textbf{0.82} & \underline{0.37} & 2.73 & \underline{3.80} & \underline{0.81} & \underline{0.48} & \underline{3.00} & \underline{3.53} \\ \midrule
\multirow{3}{*}{\makecell[l]{AEGIS \\ (Qwen 3 generator)}}
 & 0.9B & 0.84 & 0.64 & \underline{3.93} & 3.00 & 0.65 & 0.64 & \textbf{3.73} & 3.13 & 0.61 & 0.52 & 2.90 & 3.30 \\
 & 2B   & 0.85  & 0.58 & 3.50 & 3.60 & 0.65 & 0.57 & 2.73 & \textbf{4.43} & 0.62 & 0.56 & \textbf{3.10} & 3.43 \\
 & 4.3B & 0.85 & 0.60 & 3.73 & 3.76 & 0.64 & 0.58 & \underline{3.26} & 4.23 & 0.63 & 0.59 & 3.00 & \textbf{4.63} \\
\bottomrule
\end{tabular}
}
\caption{Exploratory comparison of detoxification outputs across generator backbones. AEGIS denotes the detector-plus-prompting pipeline used in the generation experiments, while LLaMA 3.2 and Qwen 3 denote the frozen generator backbones used inside the pipeline. The table is intended to analyze toxicity--meaning trade-offs rather than establish state-of-the-art performance. Bold indicates the best value and underline indicates the second-best value within each language block; LLM-S and LLM-T denote LLM-based semantic-similarity and toxicity-reduction scores.}
\label{tab:main}
\end{table*}

%% file: tables/guide_and_reflect.tex
\begin{table}[!t]
\centering
\resizebox{\columnwidth}{!}{
\begin{tabular}{l|cccc}
\toprule
Method & BERT $\uparrow$ & Tox. $\downarrow$ & LLM-S $\uparrow$ & LLM-T $\uparrow$ \\
\midrule\midrule
Unguided       & 0.82 & 0.84 & 1.87 & 4.37 \\
Guided         & \textbf{0.90} & 0.35 & 3.00 & 4.00 \\
Guided+Reflect &   0.88    &   \textbf{0.25}   & \textbf{3.19} & \textbf{4.66} \\
\bottomrule
\end{tabular}
}
\caption{Ablation of guidance and reflection strategies within the AEGIS pipeline. LLM-S and LLM-T denote LLM-based semantic-similarity and toxicity-reduction scores, respectively.}
\label{tab:methods}
\vspace{-15pt}
\end{table}

%% file: figures/abl_iteration.tex
\begin{figure*}[!t]
\centering
\begin{subfigure}[t]{0.48\textwidth}
    \centering
    \includegraphics[width=\textwidth]{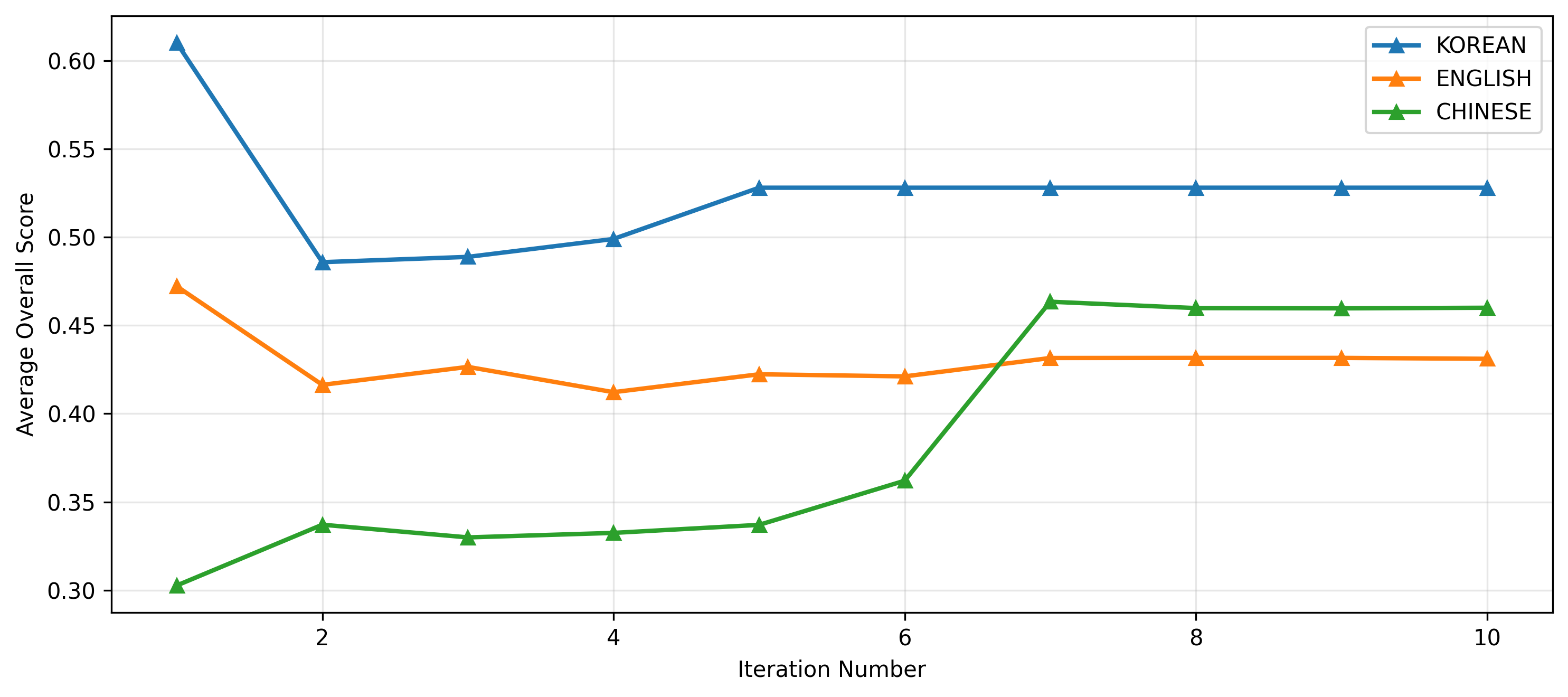}
    \caption{Overall score by iteration.}
    \label{fig:overall}
\end{subfigure}
\hfill
\begin{subfigure}[t]{0.48\textwidth}
    \centering
    \includegraphics[width=\textwidth]{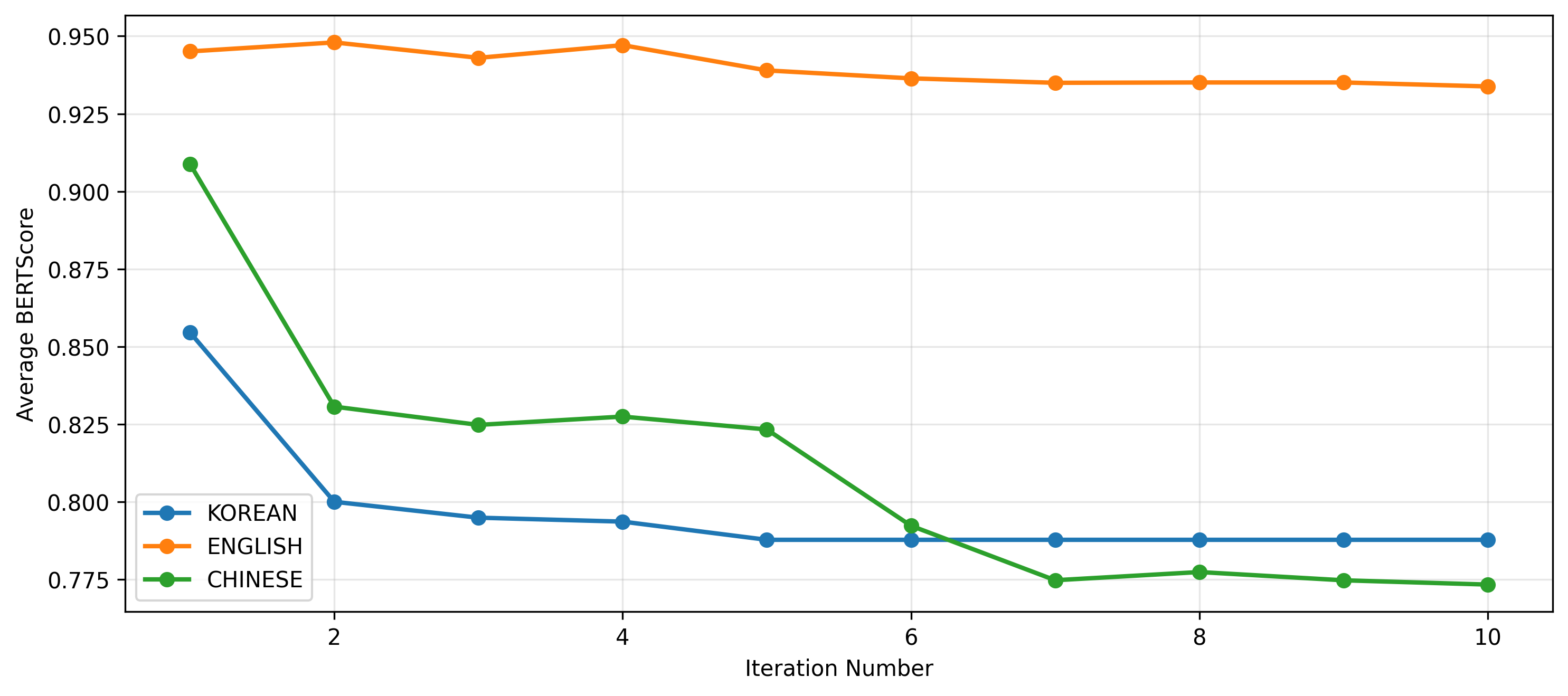}
    \caption{Semantic similarity (BERTScore) across iterations.}
    \label{fig:bertscore}
\end{subfigure}

\vspace{6pt}

\begin{subfigure}[t]{0.48\textwidth}
    \centering
    \includegraphics[width=\textwidth]{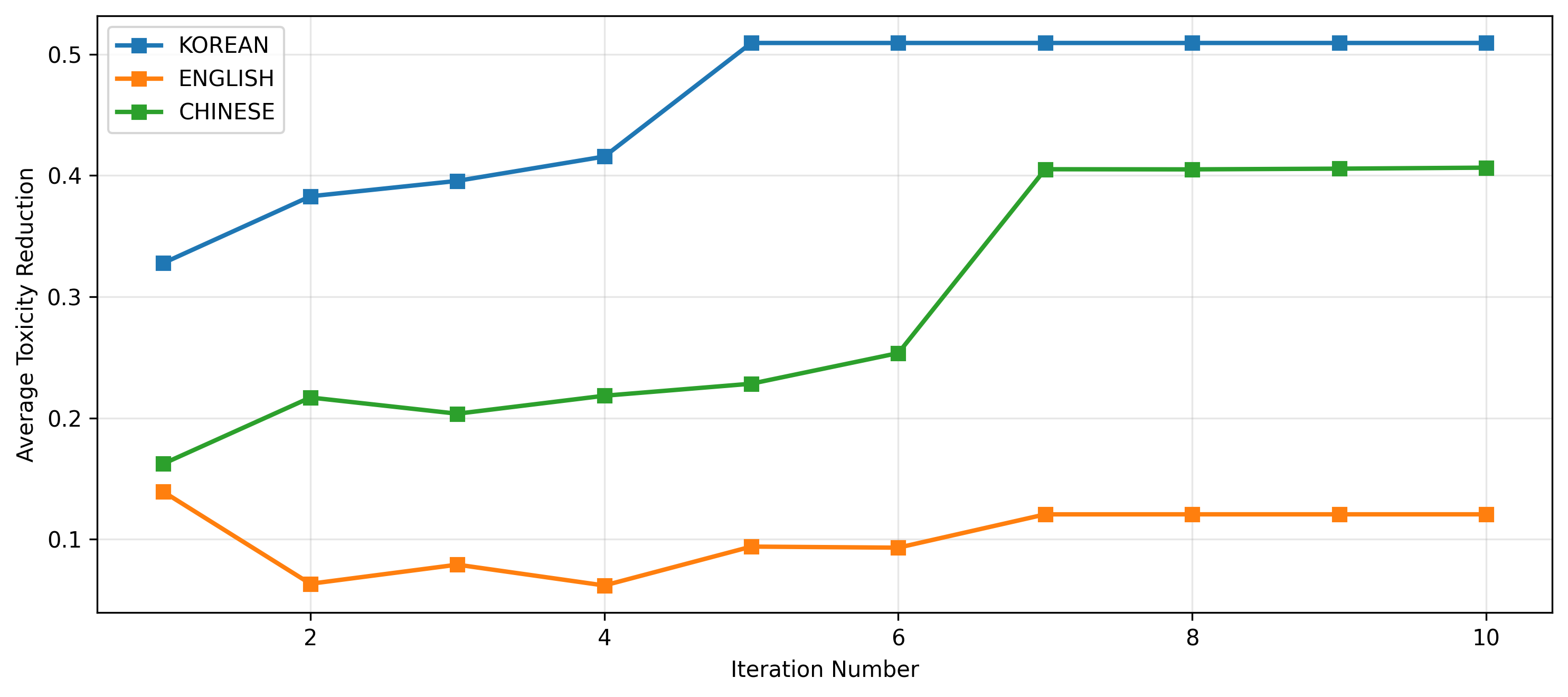}
    \caption{Toxicity reduction rate per iteration.}
    \label{fig:toxicity}
\end{subfigure}
\hfill
\begin{subfigure}[t]{0.48\textwidth}
    \centering
    \includegraphics[width=\textwidth]{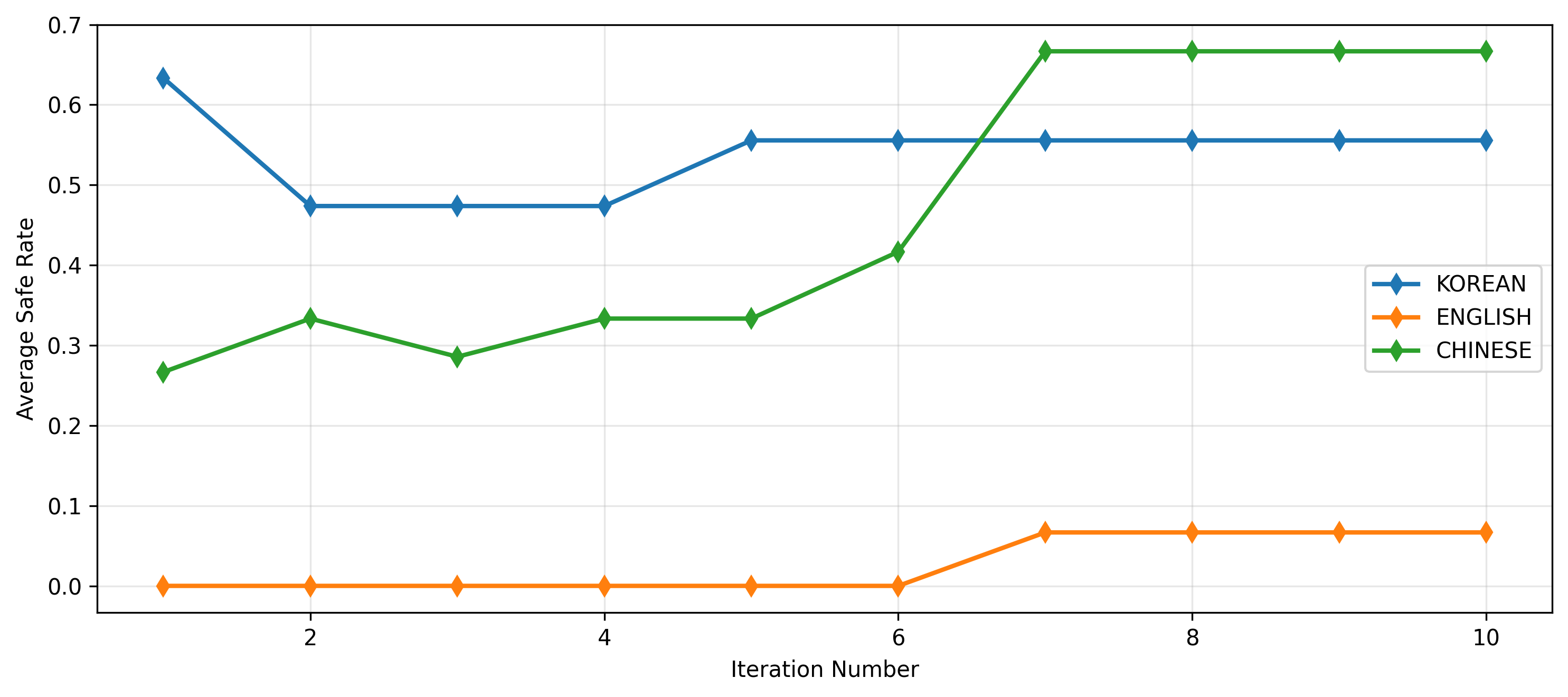}
    \caption{Detector-safe rate improvement over iterations.}
    \label{fig:safe_rate}
\end{subfigure}

\caption{
Iteration-wise behavior of AEGIS under self-refinement. The plots illustrate toxicity--meaning trade-offs across refinement steps rather than monotonic improvement on all dimensions.
}
\label{fig:detox_comparison}
\end{figure*}

%% file: tables/detector.tex
\begin{table}[t!]
\centering
\resizebox{\columnwidth}{!}{
\begin{tabular}{l|l|cccc}\toprule
\textbf{Language} & \textbf{Model} & \textbf{BIO F1} & \textbf{Non-O F1} & \textbf{Intensity F1} & \textbf{Target F1} \\ \midrule\midrule
\multirow{2}{*}{English} & XLM-R & 93.89 & \textbf{78.40} & 68.17 & 60.15 \\
 & InfoXLM & \textbf{94.05} & 74.69  & \textbf{68.24} & \textbf{61.33} \\ \midrule
\multirow{2}{*}{Korean} & XLM-R & \textbf{97.19} & \textbf{68.50} & \textbf{65.97} & 54.71 \\
 & InfoXLM & 97.01 & 66.99 & 65.48 & \textbf{55.32} \\ \midrule
\multirow{2}{*}{Chinese} & XLM-R & 95.42 & \textbf{72.26} & \textbf{78.69} & \textbf{59.50} \\
 & InfoXLM & \textbf{95.54} & 68.54 & 77.98 & 58.97 \\ \bottomrule
\end{tabular}
}
\caption{Detector validation as fixed instrumentation for span-guided generation. All metrics are denoted with \%.}
\label{tab:detector_performance}
\vspace{-7pt}
\end{table}

%% file: tables/qualitative_results.tex
\begin{table}[!t]
\centering
\resizebox{\columnwidth}{!}{
\begin{tabular}{l|l|l} \toprule
 & & \textbf{Text} \\ \midrule\midrule

\multirow{2}{*}{English}
& Original   & "You are a f**king idiot!" \\ \cmidrule{2-3}
& Detoxified & "You're being really thoughtless!" \\ \midrule\midrule

\multirow{3}{*}{Chinese}
& Original   & \makecell[l]{
\zh{"残疾人就是社会的负担"} \\
\textcolor{gray}{\small (Disabled people are a burden to society.)}
} \\ \cmidrule{2-3}
& Detoxified & \makecell[l]{
\zh{"残疾人是社会的一部分"} \\
\textcolor{gray}{\small (Disabled people are a part of society.)}
} \\ \midrule\midrule

\multirow{3}{*}{Korean}
& Original   & \makecell[l]{
"뭘 잘못 했는지도 모르네. 좌빨 놈들!" \\
\textcolor{gray}{\small (They don't even know what they did wrong. Those leftist bastards!)}
} \\ \cmidrule{2-3}
& Detoxified & \makecell[l]{
"뭘 잘못 했는지도 모르네." \\
\textcolor{gray}{\small (They don't even know what they did wrong.)}
} \\ \bottomrule

\end{tabular}
}
\caption{Examples of original and detoxified texts from AEGIS across English, Chinese, and Korean. English translations of non-English examples are shown in gray.}
\label{tab:qualitative}
\vspace{-10pt}
\end{table}